\documentclass[letterpaper,10pt,fleqn,notitlepage]{article}

\newcommand{\diag}{\mathop{\mathrm{d}}}
\newcommand{\trace}{\mathop{\mathrm{tr}}}
\newcommand{\rank}{\mathop{\mathrm{rank}}}

\newcommand{\A}{\ensuremath{\mathbf{A}}}
\newcommand{\B}{\ensuremath{\mathbf{B}}}
\newcommand{\C}{\ensuremath{\mathbf{C}}}

\newcommand{\I}{\ensuremath{\mathbf{I}}}

\newcommand{\K}{\ensuremath{\mathbf{K}}}

\newcommand{\M}{\ensuremath{\mathbf{M}}}
\newcommand{\N}{\ensuremath{\mathbf{N}}}

\newcommand{\Q}{\ensuremath{\mathbf{Q}}}
\newcommand{\RR}{\ensuremath{\mathbf{R}}}
\renewcommand{\SS}{\ensuremath{\mathbf{S}}}
\newcommand{\T}{\ensuremath{\mathbf{T}}}
\newcommand{\U}{\ensuremath{\mathbf{U}}}
\newcommand{\V}{\ensuremath{\mathbf{V}}}

\newcommand{\X}{\ensuremath{\mathbf{X}}}
\newcommand{\Y}{\ensuremath{\mathbf{Y}}}

\newcommand{\kk}{\ensuremath{\mathbf{k}}}

\renewcommand{\t}{\ensuremath{\mathbf{t}}}

\newcommand{\x}{\ensuremath{\mathbf{x}}}
\newcommand{\y}{\ensuremath{\mathbf{y}}}

\newcommand{\norm}[1]{\left\lVert#1\right\rVert}

\usepackage{amsmath, amsthm, amssymb}

\date{}

\newtheorem{lem}{Lemma}
\title{On the closed-form solution of the rotation matrix arising in computer vision problems}
\author{Andriy Myronenko and Xubo Song \\
{\it \normalsize Dept. of Science and Engineering, School of Medicine} \\
{\it \normalsize Oregon Health and Science University, Portland, OR}\\
{\normalsize homepage:} {\tt \normalsize www.csee.ogi.edu/\raisebox{-0.7ex}{\~{ }}myron}\\
{\normalsize email:} {\tt \normalsize myron@csee.ogi.edu}\\
}

\begin{document}
\maketitle

\begin{abstract}
We show the closed-form solution to the maximization of $\mathop{\mathrm{tr}}(\A^T\RR)$, where $\ensuremath{\mathbf{A}}$ is given and
$\ensuremath{\mathbf{R}}$ is unknown rotation matrix. This problem occurs in many computer vision tasks
 involving optimal rotation matrix estimation. The solution has been continuously reinvented in different
 fields as part of specific problems. We summarize the historical evolution of the problem and present
the general proof of the solution. We contribute to the proof by considering the degenerate cases of
$\ensuremath{\mathbf{A}}$ and discuss the uniqueness of $\ensuremath{\mathbf{R}}$.
\end{abstract}

\section{Introduction}
\label{sec:intro}
 Many computer vision problems that require estimation of the optimal rotation matrix reduce to 
 the maximization of $\trace(\A^T\RR)$\footnote{Matrix \emph{trace}, $\trace()$, stands for a sum of diagonal elements of the matrix. 
  $\trace(\A^T\RR)$ also represents a Frobenius inner product, which is 
 a sum of element-wise products of matrices $\A$ and $\RR$.} for a given matrix $\A$:
 \begin{equation}
 \label{eq:main}
  \text{max}\ \trace(\A^T\RR), \ \ \text{s.t.}\ \RR^T\RR=\I,\ \det(\RR)=1. 
 \end{equation}
 \noindent For instance, to estimate the closest rotation matrix $\RR$ to the given matrix $\A$,
 we can minimize:
 \begin{multline}
  \text{min}\ \norm{\RR-\A}^2_F=\trace((\RR-\A)^T(\RR-\A))=\trace(\RR^T\RR+\A^T\A)-2\trace(\A^T\RR)=\\
  \trace(\I+\A^T\A)-2\trace(\A^T\RR)= -2\trace(\A^T\RR)+\text{const.}
 \end{multline}
which is equivalent to the problem in Eq.~\ref{eq:main}. 
Historically, matrix $\RR$ was first constrained to be only orthogonal ($\det(\RR)=\pm 1$), 
which includes rotation and flip. A brief list of the optimization
problems that simplify to the maximization of $\trace(\A^T\RR)$ include:
\begin{itemize}
 \item $\text{min}\ \norm{\RR-\A}^2_F$ : the closest orthogonal approximation problem~\cite{Fan55,Higham86},\\
 \item $\text{min}\ \sum_i (\x_i-\RR \y_i)^2 = \norm{\X-\RR\Y}^2_F$ : orthogonal Procrustes problem~\cite{Schonemann66,Hanson81},
 where $\X=(\x_1,\dots,\x_N)^T,\Y=(\y_1,\dots,\y_N)^T$ are matrices whose columns are formed from the point position vectors, \\
 \item $\text{min}\ \sum_i (\x_i-(s\RR \y_i+\t))^2 = \norm{\X-(s\RR\Y+\T)}^2_F$ : Absolute orientation problem
   (generalized Procrustes problem)~\cite{Arun87,Umeyama91}, where $s$ is a scaling constant and $\t$,$\T$ are translation vector and matrix respectively,\\
 \item $\text{max}\ \trace(\RR^T\A)$: Scott and Longuet-Higgins~\cite{ScottHiggins91} correspondence estimation, where $\A$ is a 
 proximity matrix.
\end{itemize}

\section{The Lemma}

\begin{lem}
\label{lem:trace_AR}  
Let $\RR_{D\times D}$ be an unknown rotation matrix and $\A_{D\times D}$ be a known real square matrix. 
Let $\U\SS\V^T$ be a Singular Value Decomposition (SVD) of $\A$, where
$\U\U^T=\V\V^T=\I, \SS=\diag(s_i), s_1 \geq s_2 \geq,\dots,\geq s_D,\geq 0.$
Then the optimal rotation matrix $\RR$ that maximizes $\trace{(\A^T\RR)}$ is 
\begin{equation}
\label{eq:lemtoprove}
 \RR=\U\C\V^T,\ \ \text{where} \ \C=\diag(1,1,\dots,1,\det(\U\V^T)). 
\end{equation}
Matrix $\RR$ is unique for any $\A$, except for two cases: 
\begin{enumerate}
 \item $\rank(\A) < D-1$,
 \item $\det(\A) < 0$ and the smallest singular value, $s_D$, is not distinct.
\end{enumerate}
\end{lem}

\section{History of the problem}

The lemma has been reinvented repeatedly in various formulations in various fields. Historically,
the problem was constrained to be only orthogonal. Here, we try to summarize the historical flow 
of the problems and its solutions that include the lemma. 

In 1952, Green~\cite{Green52} showed the solution to orthogonal Procrustes problem in the special case of the full rank positive definite $\A$,
where $\RR$ is orthogonal. In 1966, Sch{\"o}nemann~\cite{Schonemann66} generalized the Green's solution to the arbitrary $\A$ and discussed the
 uniqueness of $\RR$. In 1981, Hanson and Norris~\cite{Hanson81} presented the solution for strictly rotation matrix $\RR$.  Unfortunately, this work has not received the widespread attention.

In the context of the closest orthogonal approximation problem, similar solution has been independently found in 1955 by
Fan and Hoffman using~\emph{polar decomposition}~\cite{Fan55,Higham86}.

In 1987, Arun et al.~\cite{Arun87} presented the solution to the absolute orientation problem,
and re-derived the lemma for orthogonal $\RR$, presumably being not aware of the earlier works.
In the same year, similar to Arun's solution was independently obtained by Horn et al.~\cite{Horn87}.

In 1991, based on the Arun's work, Umeyama~\cite{Umeyama91} presented the proof for the optimal strictly rotational matrix,
once again, being not aware of Hanson and Norris, and Sch{\"o}nemann works. 
As we shall show, Umeyama did not consider all possible solutions, specifically for the degenerate cases of $\A$, 
which makes his proof slightly incomplete.

Here, we prove the lemma in general case, mainly, following the Umeyama's work~\cite{Umeyama91}. In particular, we shall also consider the degenerate cases where $\A$ has not-distinct singular values, which was only briefly mentioned by Hanson and Norris~\cite{Hanson81}, but otherwise, to our best knowledge, never considered for the estimation of the optimal rotation matrix $\RR$.

\section{Proof of the Lemma}

We convert the constrained optimization problem into unconstrained using
Lagrange multipliers. Define an objective function $f$ to be minimized as
\begin{equation}
 \text{min} \ f(\RR)= -\trace(\A^T\RR)+\trace\left((\RR^T\RR-\I)\Lambda\right)+\lambda(\det(\RR)-1),
\end{equation}
where $\Lambda$ is a symmetric matrix of unknown Lagrange multipliers and $\lambda$ is another unknown Lagrange
multiplier. Equating to zero the partial derivatives of $f$ with respect to $\RR$,
we obtain the following system of equations:
\begin{equation}
\label{eq:lebeq_deriv}
 \frac{\partial f}{\partial \RR}=-\A+\RR\Lambda+\lambda\RR= \RR\B-\A=0.
\end{equation}
where $\B$ is symmetric by construction: $\B=\Lambda+\lambda\I$.
Thus we need to solve a linear system of equations:
\begin{equation}
\label{eq:polarproblem2}
\A=\RR\B, \ \ \text{s.t.} \ \RR^T\RR=\I,\ \det(\RR)=1.
\end{equation}

Transposing Eq.~\ref{eq:polarproblem2} and multiplying from both sides we obtain:
\begin{equation}
\label{eq:polarproblem3}
\A^T\A=\B^2.
\end{equation}
The matrix $\A^T\A$ is guaranteed to be symmetric and positive definite (or semi-definite if $\A$ is singular),
and we can decompose it using spectral decomposition:
\begin{equation}
\label{eq:polarproblem4}
\B^2=\A^T\A=\V\SS^2\V^T,
\end{equation}
where $\SS^2$ is real non-negative diagonal matrix of eigenvalues of $\A^T\A$ as well as $\B^2$,
so that $s_1^2 \geq s_2^2 \geq,\dots,\geq s_D^2,\geq 0$. Also, note that the matrix $\SS$
is real non-negative diagonal matrix of the singular values of $\A$.

Clearly, matrices $\B$ and $\B^2$ are both symmetric with commutative property: $\B\B^2=\B^2\B$,
hence both share the same eigenvectors, only when $\B^2$ is not degenerative\footnote{Here by degenerative matrix we mean a matrix
with not distinct (repeated) singular values. Note, that a matrix can be non-singular, but still degenerative.}. Thus matrix $\B$ is in the form:
\begin{equation}
 \B=\V\M\V^T
\end{equation}
where  $\M$ is real diagonal matrix with eigenvalues of $\B$, which must be in the form:
 $\M=\diag(\pm s_1,\pm s_2,\dots,\pm s_D)$.

In the degenerate case of $\A$ (but still valid), $\SS$, as well as $\SS^2$, has repeated values,
and matrix $\M$ does not have to be diagonal. $\M$ has symmetric block-diagonal structure
with the number of blocks equal to the number of distinct values in $\SS^2$. To see it happening, note
that 
\begin{align}
 &\SS^2\M=\V^T\B^2\V\V^T\B\V=\V^T\B^2\B\V=\V^T\B\B^2\V=\M\SS^2, \Rightarrow \\
 &\SS^2\M-\M\SS^2=0,\Rightarrow \\
 &(s_i^2-s_j^2)m_{ij}=0,\ \forall i,j
\end{align}
where $s_i^2$, $m_{ij}$ are the elements of $\SS^2$ and $\M$ respectively. If all the $s_i^2$
are distinct, then we conclude that $m_{ij}=0,\forall i\neq j$ and $\M$ is diagonal. If not all $s_i^2$
are distinct, then $m_{ij}=0$ only if $s_i^2\neq s_j^2$, and thus $\M$ is block-diagonal formed from square 
symmetric blocks corresponding to repeated values $s_i$. 

Now, we consider the following cases separately: $\A$ is non-singular and non-degenerative (all singular values are distinct),
$\A$ is non-singular and degenerative, $\A$ is singular.

\paragraph{Non-degenerative case of $\A$:}$\M$ is diagonal. Substituting $\M$ into
 equation Eq.~\ref{eq:polarproblem2} and then into the objective function, we obtain:
\begin{equation}
 \trace(\A^T\RR)=\trace(\B^T\RR^T\RR)=\trace(\B)=\trace(\V\M\V^T)=\trace(\M)
\end{equation}
Taking into account that $\det(\RR)=1$, from Eq.~\ref{eq:polarproblem2} we see that
\begin{equation}
 \det(\A)=\det(\RR)\det(\B)=\det(\B)=\det(\V)\det(\M)\det(\V^T)=\det(\M),
\end{equation}
hence $\det(\M)$ must have at least the same sign as $\det(\A)$. Clearly, matrix $\M$ that maximizes its trace is
\begin{align}
\label{eq:lemM1}
 \M&=\diag( s_1, s_2,\dots, s_D), &if \det(\A)>0,\\
 \label{eq:lemM2}
 \M&=\diag( s_1, s_2,\dots, -s_D), &if \det(\A)<0.
\end{align}
and the value of objective function at the optimum is
\begin{equation}
\label{lem:eqvalue}
\trace(\A^T\RR)=\trace(\M)=s_1+s_2+,\dots, +s_{D-1} \pm s_{D}
\end{equation}
where the last sign depends on the determinant of $\A$.

Now, we can find the optimal rotation matrix $\RR$, from the Eq.~\ref{eq:polarproblem2}:
\begin{align}
 \A&=\RR\B,\\
 \U\SS\V^T&=\RR\V\M\V^T,\\
 \U\SS&=\RR\V\M.
\end{align} 
If $\A$ is non-singular ($\rank(\A)=D$), then $\M$ is invertable, and the optimal $\RR$ is
\begin{equation}
\label{eq:nondegener}
  \RR=\U\SS\M^{-1}\V^T=\U\C\V^T, \ \ \text{where} \ \C=\diag(1,1,\dots,1,\det(\U\V^T)).
\end{equation}
where $\det(\U\V^T)=\det(\U)\det(\V^T)=sign(\det(\A))=\pm 1$ depending on a sign of $\det(\A)$.

\paragraph{Degenerative case of $\A$ :}$\M$ is symmetric block diagonal.
Since $\M$ is symmetric, it can be diagonalized using spectral decomposition:  $\M=\Q\N\Q^T$,  where
$\Q$ is orthogonal and also block-diagonal with the same block structure as $\M$. Matrix
$\N$ is real and diagonal. 
\begin{align}
\B^2&=\V\SS^2\V^T=\V\M^2\V^T,\Rightarrow\\
\SS^2&=\M^2=\Q\N^2\Q^T, \Rightarrow\\
\N^2&=\Q^T\SS^2\Q=\SS^2.
\end{align}
The last equality holds, because $\SS^2$ has multiples of identity along the diagonal, which correspond to the
repeated values. Matrix $\Q$ is orthogonal block diagonal, where each block has a corresponding multiples of 
identity in $\SS^2$. Using direct matrix multiplication you can see that $\Q^T\SS^2\Q=\SS^2$.

Thus $\N=\diag(\pm s_1,\pm s_2,\dots,\pm s_D)$, and the value of objective function at the optimum is
$\trace(\M)=\trace(\Q\N\Q^T)=\trace(\N)$. Taking into account the sign of determinant of $\A$ we conclude that
$\N$ that maximizes its trace is in the form:
\begin{align}
\label{eq:lemN1}
 \N&=\diag( s_1, s_2,\dots, s_D), &if \det(\A)>0,\\
\label{eq:lemN2}
 \N&=\diag( s_1, s_2,\dots, -s_D), &if \det(\A)<0.
\end{align}
%We also note that if the smallest element $s_D$ is repeated $k$-times, then there is $k$ equivalent choices to constract $\N$ in Eq.~\ref{eq:lemN2}.
The objective function at the optimum is $\trace(\N)=s_1+s_2+,\dots, +s_{D-1} \pm s_{D}$,
which is exactly the same value as in Eq.~\ref{lem:eqvalue}, when $\M$ is diagonal. 
Thus, in the degenerate case there is a set of block-diagonal matrices $\M$, which give the same objective function value as for 
the diagonal $\M$.

Now, let us consider the form of $\M$ for the optimal choices of $\N$. 
When $\det(\A)>0$, from Eq.~\ref{eq:lemN1}, we have:
\begin{equation} 
  \M=\Q\N\Q^T=\Q\SS\Q^T=\SS=\N,
\end{equation}
where the orthogonal matrix $\Q$ vanishes with corresponding multiples of identity in $\SS$. Thus if $\det(\A)>0$
the optimal $\M$ is unique. In the case when $\det(\A)<0$ (Eq.~\ref{eq:lemN2}), equality $\Q\N\Q^T=\N$ holds only if the smallest element $s_D$
is not repeated. If $s_D$ happen to be repeated, and $\det(\A)<0$, then $\M$ is not unique, and there is a set optimal solutions $\M=\Q\N\Q^T$,
which is unavoidable. However, even in this case, it is always possible to choose $\M$ to be diagonal (Eq.~\ref{eq:lemM2}). 
Similar to the non-degenerative case, if $\A$ is non-singular, the optimal $\RR$ is found in the same way as in Eq.~\ref{eq:nondegener}:
\begin{equation}
\label{eq:degener}
  \RR=\U\SS\M^{-1}\V^T=\U\C\V^T, \ \ \text{where} \ \C=\diag(1,1,\dots,1,\det(\U\V^T)).
\end{equation}

However, consider the uniqueness of SVD of $\A$ and computation of $\RR$ in the degenerative case. We know that if the singular values of 
$\A$ are not distinct, then SVD of $\A$ is not unique. In particular, any normalized linear combination of singular vectors corresponding to the same singular value is also a singular vector. Consider SVD of degenerative $\A$:
\begin{align}
 \A&=\U\SS\V^T,\\
 \U&=\A\V\SS^{-1}
\end{align}
Accordingly to Eq.~\ref{eq:degener}:
\begin{equation}
 \RR=\U\C\V^T=\A\V\SS^{-1}\C\V^T
\end{equation}
If $\det(\A)>0$, $\RR=\A\V\SS^{-1}\V^T$, which means that $\RR$ is unique, eventhough $\V$ is not.
This is because $\SS^{-1}$ has the same repeated elements as singular values in $\SS$, then $\V\SS^{-1}\V^T$ is 
a unique matrix, and thus $\RR$ is uniquely determinted.

If $\det(\A)<0$ and the smallest singular value is not distinct, then the rotation matrix $\RR=\A\V\SS^{-1}\C\V^T$ is not unique,
because different SVD of $\A$ produce different $\V$, and the matrix $\V\SS^{-1}\C\V^T$ is not uniquely determined.
Furthermore, even if the singular values of $\A$ are distinct but poor isolated (close to each other), 
a small perturbation to $\A$ can alter a singular vectors significantly~\cite{Golub89}, and thus $\RR$ changes significantly as well. 
This means, that in case of $\det(\A)<0$ and degenerative $\A$ or close to degenerative, matrix $\RR$ is extremely sensitive to 
any changes in $\A$. In particular, in this case, a round-off errors presented in computation of $\A$ and SVD of $\A$,
can produce significantly different $\RR$. 
 
We note, that Umeyama~\cite{Umeyama91}, in his derivation of the lemma,  has not considered the case when $\A$ is degenerative.

\paragraph{Singular case of $\A$:}If $\A$ is singular and $\rank(\A)=D-1$ (only a single singular value is zero), then $\M=\SS=\diag(s_1,s_2,\dots,0)$ and
\begin{equation}
 \U\SS=\RR\V\SS
\end{equation} 
If we define an orthogonal matrix $\K=\U^T\RR\V$, then
\begin{equation}
 \K\SS=\SS.
\end{equation}
Since the column vectors of $\K$ are orthonormal, then they are in the form
\begin{align}
\label{eq:lemK1}
 &\kk_i=(0,0,\dots,1_i,\dots,0)^T, & \text{for} \ 1\leq i\leq D-1\\
\label{eq:lemK2}
 &\kk_D=(0,0,\dots,\pm 1)^T.
\end{align}
Taking into account the constraint on determinant of $\RR$, we have
\begin{equation}
 \det(\K)=\det(\U^T)\det(\RR)\det(\V)=\det(\U\V^T)
\end{equation}
Thus, we obtain:
\begin{equation}
\label{eq:lemlast}
 \RR=\U\K\V^T=\U\C\V^T, \ \ \text{where} \ \C=\diag(1,1,\dots,1,\det(\U\V^T)).
\end{equation}
Finally, if $\A$ is singular and $\rank(\A) < D-1$  ($\A$ has multiple zero singular values),
 then matrix $\K$ is not uniquely determined. 
Precisely, one can choose arbitrary last column-vectors of $\K$,(number of which
is equivalent to the number of zero singular values) as far as they are orthonormal and  $\det(\K)=\det(\U\V^T)$.
This gives a set of equivalent solutions for $\RR$. Additional information or constraints require to make $\RR$ unique.   
We note, that it is always possible to chose $\K$ according to Eq.~\ref{eq:lemK1},\ref{eq:lemK1} and
find $\RR$ from Eq.~\ref{eq:lemlast}.
 
% However, even in this case it is always possible to chose $\K$ accordingly to Eq.~\ref{eq:lemK1}-\ref{eq:lemK2}, which is partially justified by the facts that: a) depending on a problem, matrix $\A$ is often computed with errors, and the true $\A$ is non-singular. 
% b) a small regularization to the matrix $\A$, e.g. in the form of $\epsilon \I$, also makes $\A$ non-singular. 

Thus, we have considered all cases of $\A$, which concludes the lemma. 

\section{Discussion and conclusion}

The lemma is of general interest and is usefull in many computer vision and machine learning problems
that can be simplified to maximization of $\trace(\A^T\RR)$. The lemma shows the optimal solution for 
the rotation matrix $\RR$. In most of the cases $\RR$ is uniquely determined. 
In the case when $\rank(\A)< D-1$ and in the degenerate case, 
when the smallest singular value is not distinct and $\det(\A)<0$,
the presented solution for $\RR$ is still a global optimum of the function,
but it is not unique. Also, we have shown, that in these degenerative cases, $\RR$ is extremely sensitive
to round-off errors in $\A$. In the cases when $\RR$ is not unique, the solution given by Eq.~\ref{eq:lemtoprove}
should be further justified by a particular problem. 

If we relax the constraint for $\RR$ to be strictly rotational, and allow it to be any orthogonal (which allows for rotation and flip), then
the derivation simplifies to the solution $\RR=\U\V^T$, which was established by Sch{\"o}nemann~\cite{Schonemann66},
and it is unique for all non-singular $\A$. The lemma can be applied for the problems of arbitrary dimensions.

\end{document}